\documentclass[final, 5p, times, twocolumn]{elsarticle}

\usepackage{lineno,hyperref}
\usepackage{times}
\usepackage{graphicx,epsfig,overpic,color}
\usepackage{url,amsmath,amssymb,amsthm,bm}
\usepackage{multirow,proof}
\usepackage{threeparttable,array,tabularx}

\usepackage[ruled,boxed]{algorithm2e}

\usepackage{stfloats}
\usepackage{epstopdf}

\usepackage[caption=false,font=normalsize,labelfont=sf,textfont=sf]{subfig}

\journal{X}

\biboptions{authoryear}

\begin{document}

\begin{frontmatter}

\title{Fast Measuring Pavement Crack Width by Cascading Principal Component Analysis}


\author[]{Zhicheng Wang}
\ead{ZhichengWang@emails.bjut.edu.cn}

\author[]{Junbiao Pang}
\ead{junbiao\_pang@bjut.edu.cn}


\address[]{Faculty of Information Technology, Beijing University of Technology, No.100 Pingleyuan Road, Chaoyang District, Beijing 100124, China}

\begin{abstract}

Accurate quantification of pavement crack width plays a pivotal role in assessing structural integrity and guiding maintenance interventions. However, achieving precise crack width measurements presents significant challenges due to: (1) the complex, non-uniform morphology of crack boundaries, which limits the efficacy of conventional approaches, and (2) the demand for rapid measurement capabilities from arbitrary pixel locations to facilitate comprehensive pavement condition evaluation. To overcome these limitations, this study introduces a cascaded framework integrating Principal Component Analysis (PCA) and Robust PCA (RPCA) for efficient crack width extraction from digital images. The proposed methodology comprises three sequential stages: (1) initial crack segmentation using established detection algorithms to generate a binary representation, (2) determination of the primary orientation axis for quasi-parallel cracks through PCA, and (3) extraction of the Main Propagation Axis (MPA) for irregular crack geometries using RPCA. Comprehensive evaluations were conducted across three publicly available datasets, demonstrating that the proposed approach achieves superior performance in both computational efficiency and measurement accuracy compared to existing state-of-the-art techniques.

\end{abstract}


\end{frontmatter}


\section{Introduction}

Highways serve as vital infrastructure components that significantly contribute to economic growth by facilitating nationwide connectivity and enabling access to employment opportunities, healthcare services, and other critical societal functions. Contemporary civil engineering practices have evolved from conventional ``scrap and build'' methodologies to adopt more sustainable approaches that emphasize the preservation and extension of existing road network lifespans. Nevertheless, the formation of surface cracks compromises the pavement's waterproofing integrity, permitting water infiltration into the sub-grade and consequently compromising structural stability~\cite{DUNG-fcnn-automatinon-19}. This degradation process can escalate into more severe pavement distresses, as a result of sub-grade deterioration under sustained vehicular loading conditions~\cite{ZHOU-autiomation-construction-20}.

Recent progress in image-based crack detection and width quantification has enabled the development of sophisticated methodologies capable of generating highly precise crack segmentation~\cite{zhou-tip-crack-segmentation-2019},~\cite{RandomStructuredForests-2016-YongShi},~\cite{XU-eaai-crack-segmentation-23}. While these segmentation results effectively identify crack locations, they do not inherently provide information about the structural significance of the detected defects. The assessment of crack severity depends on the analysis of key geometric parameters, including crack length, width, depth, and spatial distribution. Currently, transportation authorities continue to rely on skilled professionals to perform quantitative crack measurements. However, manual inspection techniques are constrained by their inability to comprehensively sample cracks, as measurements are typically limited to specific inspection points. Furthermore, conventional manual crack assessment approaches are associated with significant drawbacks, including being time-consuming, hazardous, labor-intensive, and susceptible to human variability~\cite{RandomStructuredForests-2016-YongShi}. These limitations collectively contribute to inaccuracies in the quantification of crack width.

\begin{figure}[t!]
\centering
\subfloat[Before Rotation]{\includegraphics[width=2.0in]{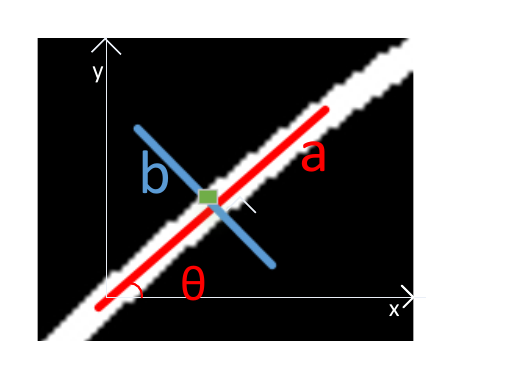}
}
\subfloat[After Rotation]{\includegraphics[width=1.6in]{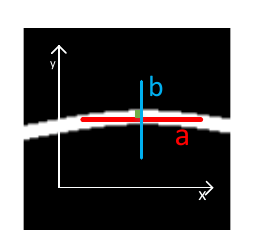}
}
\caption{A crack patch is rotated to parallel the Main Propagation Axis (MPA) of the crack. The green point is the width measurement point we selected, the red line $a$ is MPA of the crack, and the blue line $b$ is perpendicular to the red line.}
\label{fig:definition}
\end{figure}
\begin{figure}[t!]
\centering
\subfloat[]{\includegraphics[width=1.4in]{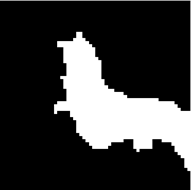}
}
\hfil
\subfloat[]{\includegraphics[width=1.4in]{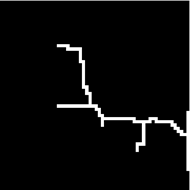}
}
\caption{(a)  is a crack and (b) is the crack skeleton corresponding to (a).}
\label{fig:skeleton}
\end{figure}
\begin{figure}[t!]
\centering
\includegraphics[width=4.4cm]{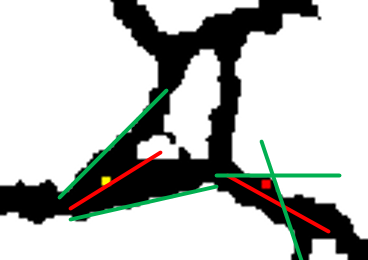}
\caption{ The yellow and red pixels are the check points we selected; the green lines are the edge lines which correspond to the check points, and the red line is the MPA (best viewed in color).}
\label{fig:pixel_wise}
\end{figure}

Over the last twenty years, image-based approaches have been successfully implemented for the automated surface surveying of cracks~\cite{dcnn-2016-zhang}.A common approach adopted by numerous researchers involves calculating the mean crack width through averaging techniques~\cite{raza-22-consruction}~\cite{XU2023real-time}. A significant limitation of these methods, however, is their failure to account for the crack's propagation direction, which consequently leads to an underestimation of the actual crack width. Specifically, Fig.~\ref{fig:definition} illustrates the inherent difficulty in unsupervisedly determining the Main Propagation Axis (MPA) of a given crack. An accurate measurement should quantify the crack width as the pixel count along a line $b$ that is oriented perpendicularly to the crack's MPA. Nevertheless, the extraction of the MPA from crack imagery is complicated by three primary challenges:
\begin{itemize}
\item \textbf{Irregular Crack with Non-Smooth Boundary:}
Fig.~\ref{fig:skeleton}(a) shows that the boundary of crack is non-smooth, and irregular. It is difficult to define the propagation main axis of crack. A naive method is to simplify the crack as a rectangle object; as a result, the width is equal to the area divides the skeleton~\cite{AutomaticDetectionChara-2012-Oliveira}. Obviously, it is not precisely describe the crack by skeleton.
Because the skeleton is affected by the non-smooth boundary, as shown in Figure~\ref{fig:skeleton}(b).

\item \textbf{Pixel-Wise Measurement:}
The shape of some crack is intricately criss-cross, as shown by the red point in Fig.~\ref{fig:pixel_wise}. The MPA is barely detected easily. Because the boundary of the crack barely is the paralleled lines. However, in principle, the crack width at every pixel should be precisely measured.
\item \textbf{Fast and Accurately Measure:} Different type of cracks have different complexity. However, precise classification of different cracks is another open problem. Therefore, a method is expected to measure the width of cracks at any check point.
\end{itemize}

In this paper, we propose a cascade method to obtain the MPA based on the following motivations: 
\begin{itemize}
    \item \emph{Easiness v.s. Difficultly:} Pavement distress is typically generalized into three types: transverse, longitudinal, and alligator cracks.We classify these into two distinct groups: low-complexity (transverse and longitudinal) and high-complexity (alligator). As illustrated in Figure~\ref{fig:pixel_wise}, low-complexity instances, such as the one indicated by the red point, exhibit geometries that approximate a rectangle. For these cases, the classical Principal Component Analysis (PCA) is sufficient to robustly identify the MPA. Conversely, for high-complexity geometries, we employ Robust PCA (RPCA).
    \item \emph{Effectiveness v.s. Efficiency:} Empirically, the occurrence of high-complexity cracks is relatively low. Therefore, PCA and RPCA are structured as a rejection cascade. This architecture is explicitly designed to balance the computational efficiency of PCA with the robust effectiveness of RPCA. In essence, the objective is to develop a computationally lightweight yet effective method capable of measuring crack width at any arbitrary check point in an unsupervised manner.
\end{itemize}

\begin{figure}[t!]
    \centering
    \includegraphics[width=0.8\linewidth]{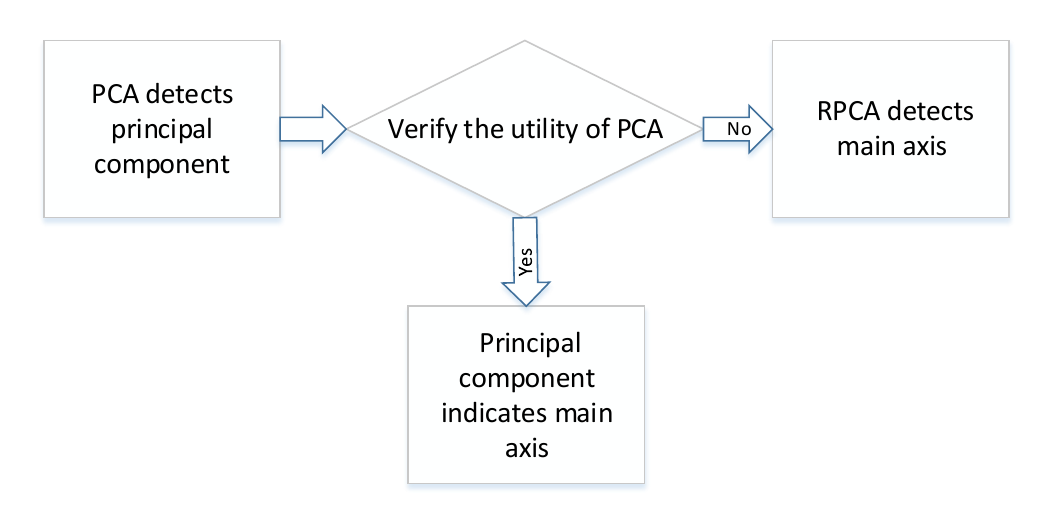}
    \caption{PCA and RPCA are organized into rejection chain}
    \label{fig:pca-rpca}
\end{figure}
In this paper, we organize PCA and Robust PCA into the Cascade PCA (CPCA) in Fig.~\ref{fig:pca-rpca}. Concretely, the traverse, and longitudinal cracks are measured by PCA, while the alligator crack are measured by RPCA. The advantage of CPCA lies to balance the speed between PCA and RPCA and obtain the high accuracy of the width measurement in an unsupervised method.
To our knowledge, this work is the first to structure PCA-based techniques into an unsupervised rejection cascade for this application. We present a comprehensive experimental evaluation to demonstrate the benefits of this novel approach for crack width measurement. The proposed method is distinguished by its computational simplicity and effectiveness. By integrating PCA, Hough voting, and RPCA within a rejection chain, our approach yields performance that is competitive with, or superior to, current state-of-the-art methods, without requiring complex parameter tuning or a training phase.


\section{Related Work}\label{sec:relatedwork}

\subsection{Pixel-Wise Crack Detection}

Algorithms for pixel-level crack detection operate by assigning a categorical label to every pixel within an image, thereby achieving a comprehensive, pixel-by-pixel detection map. This task is fundamentally an application of semantic segmentation methodologies. Notable examples include the utilization of U-Net~\cite{Ronneberger2015Biomedical} by Jenkins \textit{et al.}\cite{David2018Semantic} and the introduction of DeepCrack~\cite{Qin2018DeepCrack} specifically for this purpose. Both approaches leverage an encoder-decoder architecture, which classifies pixels as either 'crack' or 'background'. This is achieved by using multiple down-sampling stages to capture the latent crack structure and multiple up-sampling stages to facilitate feature fusion. In parallel, \cite{Yang2017Estimation} conducted an experimental evaluation on the influence of U-Net's architectural parameters—such as layer depth, convolution kernel dimensions, and feature map quantity—on both segmentation accuracy and computational speed. Their findings offer a valuable reference for optimizing network design.

Moreover, within the domain of semantic segmentation, feature fusion has been identified as a critical component for performance enhancement. This technique allows for the integration of contextual information derived from disparate feature levels, culminating in more refined and accurate crack detection outcomes~\cite{Qu2021Pavement}~\cite{Xiang2022concrete}~\cite{Song2020automated}~\cite{liu2019deepcrack}.

Nevertheless, the inherent spatial complexity of crack structures presents a significant challenge for pixel-level segmentation. The manual, pixel-by-pixel annotation required during pre-processing is highly dependent on the subjective judgment of human annotators. This annotation process is notoriously resource-intensive, being costly, laborious, and susceptible to errors such as mislabeling or omissions. From a practical standpoint, the industry places high importance on these subtle, fine-grained cracks, as their early-stage repair is more economical than addressing severe degradation. This context necessitates that any subsequent crack width measurement technique must exhibit robustness, performing reliably regardless of variations in the underlying segmentation method used to generate its input.

\subsection{Crack Width Measurement}

Methodologies for crack quantification are broadly classified as either semi-automatic or fully automatic. Semi-automatic approaches typically couple an image acquisition system with subsequent manual analysis, where operators assess the crack severity. For instance, while straightforward, techniques like the crack scale method~\cite{Yamaguchi2010Practical} are known to be time-consuming. Conversely, automatic methods depend on image processing algorithms to perform the analysis, identification, and assessment of cracks~\cite{WeixingWangA2019Pavement}. These automatic approaches can be differentiated into three primary research streams.

The first stream relies on geometric simplification, modeling a crack segment as a rectangle-like shape. For example,~\cite{liu2021crackformer} defined width as the shortest distance from the crack's skeleton to its boundaries. \cite{Benz2021Vision} adopted a method that measured width by transforming the grayscale pixel distribution of the crack into a rectangle of equivalent area. The fundamental drawback of these methods is their reliance on this rectangular assumption to identify the MPA. This simplification fails for complex geometries like alligator cracks, often resulting in an underestimation of the true width.

The second research avenue focuses on leveraging the crack's skeleton to ascertain its propagation direction. Illustratively,~\cite{Takafumi2011Concrete} applied cardinal spline interpolation to the skeleton and then utilized the derivative of the interpolated curve to determine orientation. ~\cite{Wang2017Methodology} introduced an orthogonal projection method, which measures width by simplifying the crack geometry using contours formed from boundary points. However, reliably extracting a meaningful skeleton from alligator cracks is a significant challenge for these techniques, as both spline interpolation and orthogonal projection struggle with such complex topologies.

The third research line operates on the assumption that small crack segments possess parallel boundaries, enabling width measurement based on this local geometry~\cite{Weng2019Segment}~\cite{Ni2019Zernike}. Examples include finding the Euclidean distance along a normal line intersecting the boundary pixels~\cite{Kim2019Image-based}, or determining orientation using a central difference scheme on skeleton points~\cite{Payab2019Review}. \cite{Wang2018PavementCrack} posited a direct correspondence between pixels on opposing boundaries. This "parallel boundary" hypothesis, however, rarely holds true for intricate criss-cross patterns or the complex peripheries of alligator cracks. As discussed by~\cite{Wang2017Methodology}, the presence of non-parallel boundaries or regions of high curvature can, in fact, lead to an overestimation of the crack width.

\section{Cascade PCA}\label{sec:method}

\newtheorem{myobr}{Observation}
\newtheorem{mydef}{Definition}
\newtheorem{mythe}{Theorem}
\newtheorem{mypro}{Proposition}

\subsection{Problem Definition}
\begin{mydef}[Check Point] 
A check point is defined as a specific location on the crack where the width measurement is to be performed. 
\end{mydef}

For a given check point, such as those illustrated in Fig.~\ref{fig:pixel_wise}, the challenge in width measurement is the identification of the crack's MPA. This is achieved by analyzing a local image patch $\mathbf{D}\in \mathbb{R}^{H\times W}$, where $H$ and $W$ denote the height and width of this patch, centered on the check point. The MPA is formally defined as:
\begin{mydef}[MPA]
The MPA is conceptualized as a line $y=f(x)$, parameterized by $f(x)=tx+b $, where $t\in \mathbb{R}^1$ represents the slope and $b\in \mathbb{R}^1$ is the y-intercept. For any point $p_i$ sampled from the MPA, a line perpendicular to $y=f(x)$ will intersect the crack boundaries at points $\mathbf{c}_i\in \mathbb{R}^{2\times 1}$ and $\mathbf{d}_i \in \mathbb{R}^{2\times 1}$. The optimal MPA is one that minimizes the sum of these distances, satisfying the following constraint:
\begin{equation}\label{eqt:MPA}
\ min_{t,b}\sum_{i=1}^{N} |\mathbf{p}(t,b)_i-\mathbf{c}_i|_2+ |\mathbf{p}(t,b)_i - \mathbf{d}_i|_2,
\end{equation}
where $\mathbf{p}(t,b)_i$ denotes a point on the MPA defined by parameters $t$ and $b$. Intuitively, the ideal MPA should be oriented as parallel as possible to the crack's boundaries.
\end{mydef}

If the MPA is determined, the corresponding rotation angle $\theta$ is calculated as:
\begin{equation}\label{eqt:slope}
\theta=\arctan t.
\end{equation}
The image patch $\mathbf{D}$ is then rotated by this angle $ \theta $, resulting in the orientation shown in Fig.~\ref{fig:definition}(b). This alignment allows the crack width to be measured directly as the pixel count along the vertical $y$-axis. However, finding a direct solution for Eqt.~\ref{eqt:MPA} is computationally challenging.

In this paper, we propose to use the transformation of the image patch into a low-rank matrix as a surrogate objective for Eqt.~\eqref{eqt:MPA}. Table \ref{fig:Symbolic} summarizes the key notations used throughout this paper.

\begin{table}[t!]
\centering
\begin{tabular}{|c|c|}
\hline
Notation & Definition  \\\hline\hline
$\mathbf{I}$ & Binary Crack Image \\\hline
$\mathbf{D} $ &Image blocks cropped from image $\mathbf{I}$ \\\hline
$t_1$&Slope from PCA \\ \hline
$t_2$&Slope fitted by RANSAC\\ \hline
$t_3$ & Slope from RPCA \\ \hline
$\mathcal{M}$ & Boundary coordinate set\\ \hline
$\mathcal{W}$&Crack width set\\ \hline
\end{tabular}
\caption{Some important notations used in this paper.}
\label{fig:Symbolic}
\end{table}

\subsection{PCA for the Low Complexity Crack}
Given the coordinates of the pixels on the boundary of the crack $\{(x_1,y_1),\ldots, (x_N,y_N)\}$, we organize these points $(x_i,y_i), (1\leq i \leq N)$
into the matrix $\mathbf{M} \in \mathbb{R}^{2 \times N}$, where ${(x_1,x_2,\ldots,x_N)}^T$ is the first row of $\mathbf{M}$, and ${(y_1,y_2,\ldots, y_N)}^T$ is the second column of $\mathbf{M}$. The matrix  $\mathbf{M} $ is subsequently mean-centered to produce $\bar{\mathbf{M}}$.
By decomposing the covariance matrix $\bar{\mathbf{M}}\bar{\mathbf{M}}^T $, the covariance matrix and the corresponding eigenvector $\mathbf{r}\in \mathbb{R}^{1\times 2}$ are as follows:
\begin{equation}\label{eqt:PCA}
\begin{aligned}
  \max_{R}\quad &  \text{Tr}( \mathbf{r}^{T}\bar{\mathbf{M}}\bar{\mathbf{M}}^T\mathbf{r} )\\
  \mbox{s.t.} \quad &  \mathbf{r}^{T} \mathbf{r} =1.
\end{aligned}
\end{equation}
PCA~\cite{PCA_1901_PearsonK} transforms an image patch into a set of linearly orthogonal representations. The eigenvector of PCA which corresponds to the largest eigenvalue represents the MPA of the crack. Therefore, the slop of the MAP $t_1$ from PCA is obtained as follows:
\begin{equation}\label{eqt:t1}
t_1= \mathbf{r}(1),
\end{equation}
where $\mathbf{r}(1)$ is the largest eigenvector of the matrix $\bar{\mathbf{M}}\bar{\mathbf{M}}^T$.  

\begin{figure}[t!]
\centering
\includegraphics[width=1.5in]{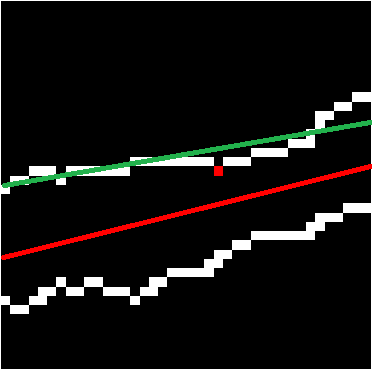}
\caption{The green line fitted by RANSAC almost parallels to the MPA of a crack (in red). The red pixel is the check pixel (best viewed in color).}
\label{fig:boundary}
\end{figure}

As illustrated in Fig.~\ref{fig:boundary}, the MPA derived from Eqt.~\eqref{eqt:t1} generally aligns well with the crack boundary. However, this alignment is predicated on a quasi-linear boundary assumption. When a crack exhibits a complex morphology, this assumption is no longer valid, causing the MPA to diverge from the boundary. Consequently, PCA proves unreliable for handling these high-complexity cases.

To address this limitation, this paper introduces a rejection chain to assess the utility of the PCA-derived result. Specifically, we employ the RANdom SAmple Consensus (RANSAC) algorithm~\cite{Ransac_1981_Fischler} as a validation mechanism. RANSAC~\cite{Ransac_1981_Fischler} is an iterative method adept at estimating mathematical model parameters from datasets contaminated with outliers. In this application, we utilize RANSAC to robustly fit a line to the crack boundary points within the image patch $\mathbf{D}$. The algorithm is thereby expected to identify and model one of the primary crack edges, disregarding the noise from boundary irregularities. Figure~\ref{fig:boundary} provides a visual verification that the line fitted by RANSAC nearly parallels to one side of the crack's boundary.

Given the slope $t_2$ of the line fitted by RANSAC, if the difference between $t_1$ and $t_2$ is less than a predefined threshold $\gamma$ as follows:
\begin{equation}\label{eqt:cascade-one}
  IslowComp(t_1,t_2, \gamma) =
  \left\{ 
  \begin{array}{cl}
    1 &  \text{if} \  \  \|t_1-t_2\|_2 \leq \gamma, \\
    0 & \text{otherwise}, 
  \end{array}
  \right.
\end{equation}
where function $IslowComp(t_1,t_2,\gamma)$ verifies whether the image block $\mathbf{D}$ has a low complexity crack or not.

\subsection{RPCA for High-Complexity Cracks}
For an image patch $\mathbf{D}$ identified as having high complexity, it is modeled as a rotated, low-rank structure corrupted by sparse noise. The underlying low-rank matrix is then leveraged to detect the MPA. Concretely, assuming the patch represents a low-rank texture $\mathbf{D}^0 \in \mathbb{R}^{H\times W}$ on a planar surface, we follow the decomposition approach of~\cite{TITL-2012-Zhang}. The observed patch $\mathbf{D}$, under a geometric transformation $\bm{\tau}$, is separated into its low-rank and sparse components as follows:
\begin{equation}\label{eqt:decompostion}
  \mathbf{D}\circ \bm{\tau}= \mathbf{D}^0+\mathbf{E},
\end{equation}
where $D^0$ is low-rank matrix and $E$ is the noise matrix. The decomposition in Eq.~\ref{eqt:decompostion} is equivalent to solving the following problem:
\begin{equation}\label{eqt:lowrank}
\begin{aligned}
  \min_{D^0,E}\quad  &  \|D^0\|_* +\lambda\left\|E\right\|_0  \\
  \mbox{s.t.} \quad & D\circ\bm{\tau} =D^0+E  
\end{aligned}
\end{equation}
where $\left\|\cdot\right\|_0 $ is the zero norm of a matrix and $\lambda$ is a weight parameter, in which
the rotation matrix is expressed by $\bm{\tau}\in \mathbb{R}^{3\times 3}$:
\begin{equation}
 \bm{\tau}=
   \begin{pmatrix}
   \cos(\theta) & \sin(\theta) & 0 \\
  -\sin(\theta) & \cos(\theta) & 0 \\
   0 & 0 & 1
  \end{pmatrix},
\end{equation}
where $\theta$ is the rotation degree.

We note that although the objective function in the above problem is convex, the constraint $D\circ\tau = D^0+E $ makes the problem difficult.
A common technique to overcome this difficulty is to locally linearize the constraint around the current estimate and iterate~\cite{AunifyingFramework-2004-Baker}. Specially, the constraint of the linearized version is as follows:
\begin{equation}\label{eqt:d}
 D\circ\bm{\tau} +\nabla D	\Delta \bm{\tau} =D^0 +E
\end{equation}
where $\nabla \mathbf{D} $ is the Jacobian: derivatives of the image \emph{w.r.t} the transformation parameters $\bm{\tau}$ and $ \Delta \bm{\tau}$ fine-tuned angle matrix.

 ADMM method is a class of algorithms that simultaneously minimize the augmented Lagrangian function and compute an
appropriate Lagrange multiplier for Eqt.~\eqref{eqt:d}.

Once the parameter $\tau$ and $\Delta \tau$ are solved, the rotation matrix of the crack image $D$ is computed as follows:
\begin{equation}\label{eqt:t}
\theta_3= \arccos \bm{\tau}_{11} + \arccos \Delta \bm{\tau}_{11},
\end{equation}
where $\bm{\tau}_{11}$ and $\Delta \bm{\tau}{11} $ are the respective $\cos(\alpha)$ elements from the rotation matrices $\bm{\tau} $ and $\Delta \bm{\tau}$. Consequently, the original problem in Eqt.~\eqref{eqt:lowrank} is simplified into the following linearized version:
\begin{equation}\label{eqt:linear-lowrank}
\begin{aligned}
  \min_{D^0,E}\quad  &  \|D^0\|_* +\lambda\left\|E\right\|_0  \\
  \mbox{s.t.} \quad & D\circ\bm{\tau} +\nabla D	\Delta \bm{\tau} =D^0 +E.
\end{aligned}
\end{equation}
This linearized problem (Eq.~\ref{eqt:linear-lowrank}) is convex and amenable to an efficient solution. Since the linearization is only a local approximation, we solve it iteratively to converge to a (local) minimum of the original non-convex problem. We optimize this objective function using the Alternating Direction Method of Multipliers (ADMM).

\begin{equation}\label{eq:lagarian-multiplier-objective}
\begin{split}
L & =\lambda \left\|\mathbf{E}\right\|_0 + \|\mathbf{D}^0\|_{*} 
 +\langle \bm{Y}, \mathbf{D} \circ \bm{\tau}+ \bm\Delta \mathbf{D} \bm\nabla \bm{\tau}-\mathbf{D}^0 - \mathbf{E}\rangle \\
& +\mu / 2\left\|\mathbf{D} \circ \bm{\tau}+ \bm\Delta \mathbf{D} \bm\nabla \bm{\tau}-\mathbf{D}^0 - \mathbf{E} \right\|_F^2
\end{split}
\end{equation}
where  $\bm{Y}$ represents the Lagrange multipliers, $\langle \bm{A,B}\rangle$ is the inner product of matrices $\bm{A}$ and $\bm{B}$, and $ \mu \geq 0$ is the penalty coefficient.  ADMM algorithms minimize this augmented Lagrangian function while concurrently computing the appropriate multipliers. Further optimization details can be found in~\cite{Zhang2010TILT}.

In practice, to accelerate the discovery of the low-rank matrix, the image patch $D$ is first pre-rotated by a discrete series of angles (e.g., $[5, 10,\ldots, 90]$). The orientation that yields the lowest-rank representation is then recorded and used as the initial $\mathbf{D}\circ \bm{\tau}$ for the optimization.

\begin{figure}[t!]
\centering
\subfloat[]{\includegraphics[width=1.5in]{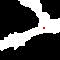}}
\hfil
\subfloat[]{\includegraphics[width=1.5in]{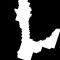}}
\hfil
\subfloat[]{\includegraphics[width=1.5in]{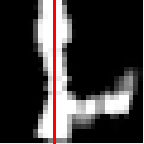}}
\hfil
\subfloat[]{\includegraphics[width=1.5in]{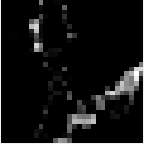}}
\caption{(a) is the image patch $\mathbf{D}$ and the red point is check point; (b) is the image patch rotated with $\bm{\tau}$; (c) is the low-rank image patch $\mathbf{D}^0$ and the red line is MPA which is across the check point; (d) is the residual image patch $\mathbf{E}$. (best viewed in color)}
\label{fig:RPCA}
\end{figure}

Figure~\ref{fig:RPCA} vividly shows how RPCA find the main axis of a crack with high complexity.

\subsection{Group PCA and RPCA into A Rejection Chain}

To balance the speed between PCA and RPCA, we group PCA and RPCA into a rejection chain as follows:
\begin{itemize}
  \item [1.]Initially, the raw crack image $\mathbf{I}$ undergoes segmentation using an established algorithm, such as CrackForest~\cite{RandomStructuredForests-2016-YongShi}. This stage yields a binary mask $\mathbf{Q}$, where pixel intensities of 255 and 0 represent the crack and background, respectively.
  \item [2.] For a given check point selected at random, a local image patch $\mathbf{D}\in \mathbb{R}^{H \times W}$ is extracted from $\mathbf{Q}$. The aspect ratio $H/W$ of this patch is determined  by a predefined threshold.
  \item [3.] Firstly, find the MAP of the crack in the image patch $\mathbf{D}$ by PCA, and the fitted line. If the $\| t_1 - t_2\|_2$ is smaller than the value $th$, the MPA of the crack at the check point is correct. Otherwise, the MPA of the image patch $\mathbf{D}$ is computed by RPCA.
\end{itemize}

\section{Experiments}\label{sec:exp}

This section details the experimental validation of our proposed method. All algorithms were implemented in Matlab and executed on a 3.30 GHz machine equipped with 6G RAM.

\begin{figure}[t!]
\centering
\subfloat[transverse]{\includegraphics[width=1.5in]{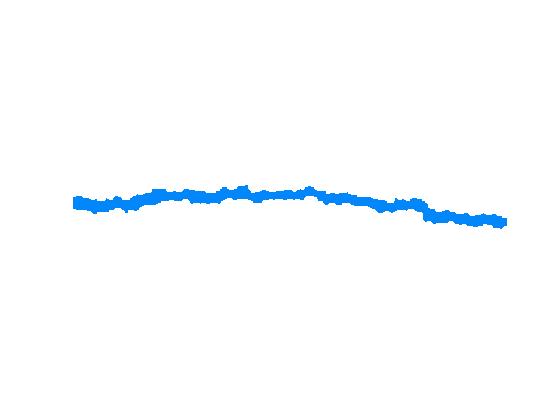}}
\hfil
\subfloat[longitudinal]{\includegraphics[width=1.5in]{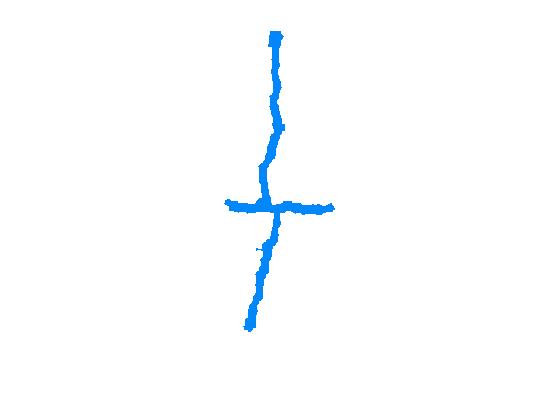}}
\hfil
\subfloat[longitudinal]{\includegraphics[width=1.5in]{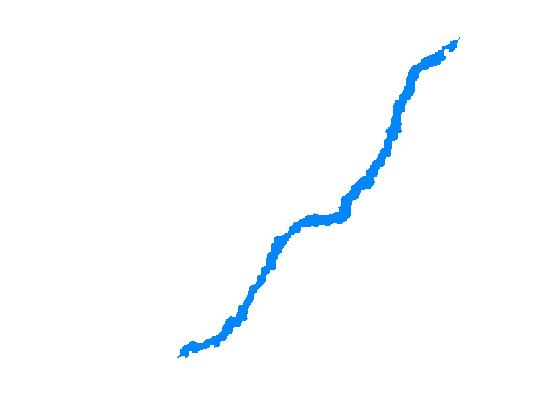}}
\hfil
\subfloat[alligator]{\includegraphics[width=1.5in]{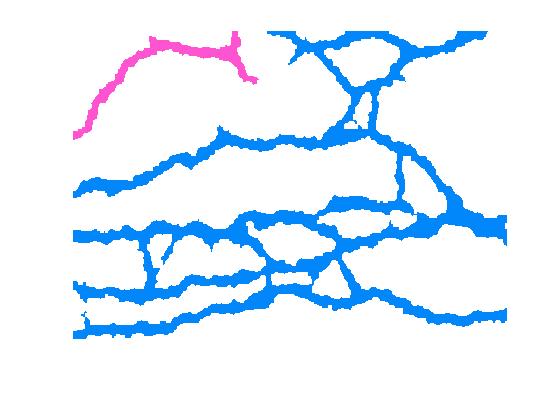}}
\caption{Some samples of crack images from CFD.}
\label{fig:example}
\end{figure}

\noindent\subsection{Database}
We validated our approach on three public datasets:

\textbf{CFD:} The CFD~\cite{shi2016automatic} dataset features cement road cracks. It comprises 118 images ($480\times320$ pixels) captured with an iPhone 5 on pavements in Beijing, China. The reported crack widths vary from 1 to 3 mm. For our ground truth, we manually annotated 8 representative images (1 horizontal, 1 longitudinal, 6 alligator). From each image, 13 points were randomly selected and labeled, yielding a total of 104 check points for CFD.
\textbf{Cracktree:} The Cracktree dataset is for the cement road crack~\cite{zou2012cracktree}. The dataset consists of $206$ crack images with a size of $800\times600$. Crack500~\cite{yan-crack500-its-2020} with 500 samples collected by cell phones. We manually label the crack width of 8 representative images (\emph{i.e.}, 1 horizontal crack, 1 longitudinal crack, 6 alligator cracks). 13 points from each image are randomly selected and labeled. Consequently, 104 check points are measured for Cracktree. 

\textbf{Crack500:} Crack500~\cite{yan-crack500-its-2020} with 500 samples collected by cell phones. We manually label the crack width of 8 representative images (\emph{i.e.}, 1 horizontal crack, 1 longitudinal crack, 6 alligator cracks). 13 points from each image are randomly selected and labeled. Consequently, 104 check points are measured for Crack500. 

To ensure an unbiased ground truth, we employed a consensus-based protocol. Five annotators were independently tasked to label each check point using the Labelme~\cite{} tool. The final ground-truth value for each point was then determined by computing the mean of these five annotations.

\subsection{Evaluation}

Performance was assessed using two standard metrics: Mean Absolute Error (MAE) and Mean Square Error (MSE). They are defined as follows:
\begin{eqnarray}
  MAE &= &\frac 1N \sum_{i=1}^N \left|w_i-gt_i\right|, \\
 MSE & = & \frac 1N \sum_{i=1}^N(w_i-gt_i)^2,
\end{eqnarray}
where $N$ is the total number of check points, $w_i$ is the measured width, and $gt_i$ is the ground-truth width for the $i$-th check point.

\subsection{Baselines and The State-of-Art Approaches}

To demonstrate the superiority of the proposed method, we compare our CPCA with the SOTA methods on these public datasets from three aspects as follows:
\begin{itemize}
\item \textbf{Combined with different Crack segmentation methods}: Two off-the-shelf crack segmentation methods including Cracktree~\cite{liu2021swin} and DeepCrack~\cite{zhou-tip-crack-segmentation-2019}, are used as baseline segmentation methods. Specially, Cracktree, based on random forest, belongs to the shallow learning approach. On the contrary, DeepCrack, based on the encoder-decoder structure, belongs to the deep learning method. We use both methods as baselines to extract pixel-wise cracks to verify the robustness of our method across different inputs.

\item \textbf{The SOTA Crack Width Measurement Methods}: The SOTA methods includes the classical Skeleton-Based Method (SBM)~\cite{skeletons-1969-Hilditch}, the combination of PCA and the nearest Point (PCAP)~\cite{Ong2022hybridmethod}, and Micro Element Method (MEM)~\cite{XU2023real-time}. The reasons that we consider them as the SOTA methods are as follows:
\textbf{SBM:} This work assumes that a segment of the crack is the rectangle-like shape. This unreasonable assumption is widely accepted to simplify the crack width measurement. However, this method ignores the jaw-saw-like shape of cracks, and the alligator crack, where the complex shape would make the assumption extremely inefficient in practice.

\textbf{PCAP:} PCAP assumes that a segment of the crack is a strip-like shape, which relaxes the assumption in SBM. Although PCAP perfectly solves the crack width measurement for the traverse and the longitudinal cracks, PCAP ignores the alligator crack. Because the MPA of the alligator barely be discovered by PCA as discovered in this paper.

\textbf{MEM:} MEM assumes that the boundary of a crack could be fitted by a line. This assumption barely holds for the alligator crack. Because the Hough voting is still sensitive to intensive noise~\cite{Jihoon2024Robust}, which is also observed in our paper.  
For a fair comparison, the hyper-parameters of these SOTA methods were tuned to achieve the best performance.
\end{itemize}

\begin{table}[t!]
\centering
\footnotesize{
\caption{Comparisons between the SOTA methods and our method by MAE and MSE on CFD.In this paper, $\underline{ ~~~~ }$ denotes the second best performance.}\label{tab:result_cfd}
\begin{tabular}{|c|c|c|c||c|c|}
\hline
\multirow{2}*{Method} &  \multirow{2}*{SBM} &\multirow{2}*{PCAP} & \multirow{2}*{MEM} &   \multicolumn{2}{c|}{Our method} \cr\cline{5-6}
& & &                   & Cracktree & DeepCrack   \cr \hline\hline
MAE & 7.2 & 4.34   &  5.30  &  $ \underline{0.98}$  & \textbf{0.91}     \cr\hline
MSE & 60.02 & 30.15 & 50.72&   $ \underline{3.15} $ &  \textbf{2.84}  \cr \hline
\end{tabular}
}
\end{table}

\subsection{Comparisons with the SOTA methods}

\textbf{Results on CFD:} Table~\ref{tab:result_cfd} shows that our method outperforms the other SOTA methods in terms of both MAE and MSE. For instance, compared with SBE, PCAP, and MEM, our method achieves 0.91 MAE, which has 6.29, 3.43, and 4.39 gains over SOTA methods, respectively. We notice that the both PCAP and MEM barely handled the alligator crack. As expected, the MSE of our method also achieves the best performances than SBE, PCAP, and MEM in Table~\ref{tab:result_cfd}. As a comparison, our method achieves competitive performances over two evaluation criteria from both MAE and MSE metrics.

Another important observation is that for the crack width measurement task, the deep learning based segmentation ( DeepCrack) minimally surpasses the the shallow based (Cracktree) on in terms of MAE and MSE, \textit{i.e.}, $0.91 vs. 0.98$ and $2.84 vs. 3.15$. The exhalation is that the samples from CFD is the cement crack that is easier than the asphalt one for crack detection~\cite{DBLP:journals-pang-its-2024}. The results indicate that our method has a good generalization ability cross different datasets.

The poor performance of the baselines confirms that their underlying assumptions (rectangle-like or strip-like) are flawed. These models are invalid for the complex boundaries of alligator cracks, where PCA alone is insufficient to discover the correct MPA.

\begin{table}[h!]
\centering
\footnotesize{
\caption{Comparisons between the SOTA methods and our method by MAE and MSE on Cracktree.}\label{tab:result_cracktree}
\begin{tabular}{|c|c|c|c||c|c|}
\hline
\multirow{2}*{Method} &  \multirow{2}*{SBM} &\multirow{2}*{PCAP} & \multirow{2}*{MEM} &   \multicolumn{2}{c|}{Our method} \cr\cline{5-6}
& & &                   & Cracktree & DeepCrack   \cr \hline\hline
MAE & 9.2 & 5.83   &  7.62  &  $ \underline{1.23}$  & \textbf{1.12}     \cr\hline
MSE & 78.02 & 43.23 & 65.35&   $ \underline{3.67} $ &  \textbf{3.15}  \cr \hline
\end{tabular}
}
\end{table}

\textbf{Results on CrackTree:} This dataset introduces greater complexity, including shadows and blurred backgrounds.

Tab.~\ref{tab:result_cracktree} shows our method consistently achieves the best performance with the 1.12 MAE and 3.15 MSE. In addition, PCAP also achieved 5.83 MAE and 43.23 MSE on Cracktree dataset. However, compared to CFD dataset, the PCA in PCAP is unable to find the MPA of the alligator cracks, resulting in a significant increase of MAE and MSE, \textit{i.e.}, 5.83-4.33 = 1.5. As a comparison, our method consistently outperforms the SOTA methods, and obtain the good generalization cross different datasets, \textit{i.e.}, 1.12-0.91=0.21.  This highlights the advantage of the PCA-RPCA combination for robustly handling diverse crack types.

\begin{table}[h!]
\centering
\footnotesize{
\caption{Comparisons between the SOTA methods and our method by MAE and MSE on Crack500.}\label{tab:result_crack500}
\begin{tabular}{|c|c|c|c||c|c|}
\hline
\multirow{2}*{Method} &  \multirow{2}*{SBM} &\multirow{2}*{PCAP} & \multirow{2}*{MEM} &   \multicolumn{2}{c|}{Our method} \cr\cline{5-6}
& & &                   & Cracktree & DeepCrack   \cr \hline\hline
MAE & 9.2 & 2.63   &  5.62  &  $ \underline{0.83}$  & \textbf{0.78}     \cr\hline
MSE & 78.02 & 28.53 & 42.35&   $ \underline{2.97} $ &  \textbf{2.83}  \cr \hline
\end{tabular}
}
\end{table}
\textbf{Results on Crack500:} Tab.~\ref{tab:result_crack500} shows the comparison results in terms of MAE and MSE on Crack500. There are two observations in Tab.~\ref{tab:result_crack500} as follows:
\begin{itemize}
\item Our method again achieves the best performance ($0.78$ MAE and $2.83$ MSE), significantly outperforming all other methods. In comparison, the MAE scores for SBM (9.2), PCAP (2.63), and MEM (5.62) are all substantially higher.
\item The MSE of our method ($2.83$) is an order of magnitude lower than that of PCAP ($28.53$). This large discrepancy in MSE underscores PCAP's critical failure in handling alligator cracks. Its inability to find the correct MPA in these cases leads to large squared errors, a problem our CPCA framework solves by invoking RPCA.
\end{itemize}

\subsection{Running Time} 

\begin{table}[h!]
    \centering
    \begin{tabular}{|c|c|c|}
    \hline
      Evaluation   & PCA & RPCA\\ \hline \hline
       Counts  & 81 & 23\\ \hline
         Time(seconds) & 0.09 & 50\\ \hline
    \end{tabular}
    \caption{The computation cost of the proposed method.}
    \label{tab:computiation-cost}
\end{table}

\begin{figure}[t!]
\centering
\subfloat[Transverse]{\includegraphics[width=1.3in]{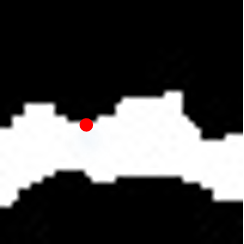}}
\hfil
\subfloat[MPA of (a)]{\includegraphics[width=1.3in]{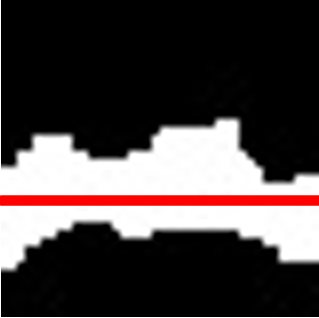}}
\hfil
\subfloat[Longitudinal]{\includegraphics[width=1.3in]{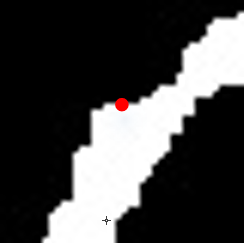}}
\hfil
\subfloat[MPA of (c)]{\includegraphics[width=1.3in]{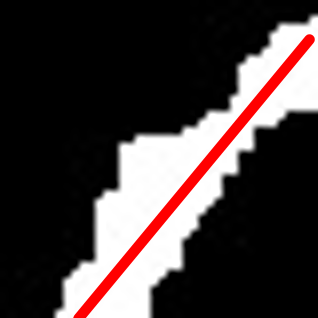}}
\hfil
\subfloat[Alligator]{\includegraphics[width=1.3in]{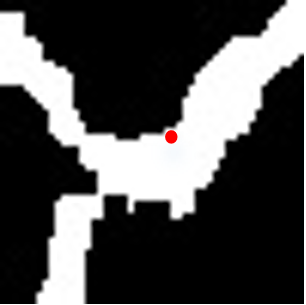}}
\hfil
\subfloat[MPA of (e)]{\includegraphics[width=1.3in]{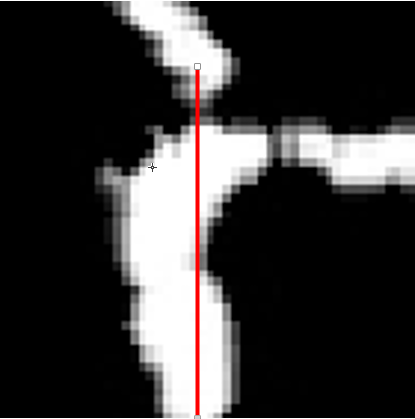}}
\hfil
\subfloat[Alligator]{\includegraphics[width=1.3in]{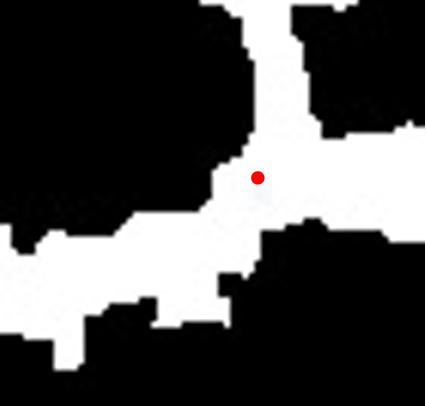}}
\hfil
\subfloat[MPA of (g)]{\includegraphics[width=1.5in]{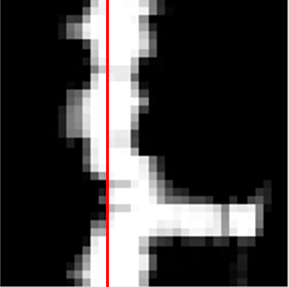}}
\caption{Visualization of our methods on CFD. Please zoom out for more details. The white indicates the cracks, the black represents the background, the red point is the check point for the width measurement, and the red line is the learned MPA (best viewed in color).}
\label{fig:resultexample}
\end{figure}

To analyze the practical efficiency of the rejection chain, we profiled the CPCA's runtime (Tab.~\ref{tab:computiation-cost}). This analysis yields two primary insights: 
\begin{itemize}
    \item \textbf{Hough voting is not always successful to find the MPA for the alligator crack.} For instance, 23 points are handled by RPCA. It verifies that the assumptions in PCAP and MEM tend to be invalid for the alligator crack. Therefore, the introduction of RPCA is necessary to find MPA for the complex shape of cracks. 
    \item \textbf{The proposed CPCA balances between accuracy and effectiveness.} only $23/104=20\%$ points are handled by the time-costing RPCA. Consequently, the rejection chain in the proposed method balances well between the running speed and accuracy.   
\end{itemize}

\subsection{Visualization} 

Figure~\ref{fig:resultexample} provides a qualitative visualization of our model's predictions on samples from CFD. For simple transverse (a) and longitudinal (c) cracks, the learned MPA correctly aligns with the crack's propagation direction, demonstrating excellent performance. More importantly, for complex alligator cracks (e, g), our method remains robust. Even when measuring at challenging intersections, the framework correctly identifies the patch as high-complexity and utilizes RPCA to find the true MPA.

\section{Conclusion}\label{sec:conclusion}

In this paper, we have described a method to efficiently measure the crack width by organizing PCA and RPCA into a cascade structure, leading to the results significantly outperforming the SOTA methods. More importantly, the proposed cascade PCA is scalable to measure the crack width from any pre-selected check points. There are significant distinctions between the proposed method and the previous studies as follows:
\begin{itemize}
\item To our best knowledge, we first notice that PCA tends to fail for the crack with the high complexity. This motivates us introduce RPCA to detect the MPA of the crack with the high complexity boundaries.    
\item We demonstrate the advantages by organizing PCA and RPCA into a rejection chain. This not only guarantees the accuracy of the proposed method, but also significantly reduces the time costs of the detection system.
\end{itemize}
The promising results of this paper motivate a further examination of the proposed CPCA. Firstly, more cracks pre-processed by different methods are used to evaluate the generalization ability of our method. Secondly, reducing the time-complexity by more advanced optimization methods should be investigated.


\section*{References}
\bibliography{acmart}

\begin{thebibliography}{41}
\expandafter\ifx\csname natexlab\endcsname\relax\def\natexlab#1{#1}\fi
\providecommand{\url}[1]{\texttt{#1}}
\providecommand{\href}[2]{#2}
\providecommand{\path}[1]{#1}
\providecommand{\DOIprefix}{doi:}
\providecommand{\ArXivprefix}{arXiv:}
\providecommand{\URLprefix}{URL: }
\providecommand{\Pubmedprefix}{pmid:}
\providecommand{\doi}[1]{\href{http://dx.doi.org/#1}{\path{#1}}}
\providecommand{\Pubmed}[1]{\href{pmid:#1}{\path{#1}}}
\providecommand{\bibinfo}[2]{#2}
\ifx\xfnm\relax \def\xfnm[#1]{\unskip,\space#1}\fi
\bibitem[{A \& C(1981)}]{Ransac_1981_Fischler}
\bibinfo{author}{A, F.~M.}, \& \bibinfo{author}{C, B.~R.} (\bibinfo{year}{1981}).
\newblock \bibinfo{title}{Random sample consensus: a paradigm for model fitting with applications to image analysis and automated cartography}.
\newblock {\it \bibinfo{journal}{Communications of the ACM}\/},  {\it \bibinfo{volume}{24}\/}, \bibinfo{pages}{381--395}.
\bibitem[{B et~al.(2019)B, A, A, A, B, C, D, E \& E}]{WeixingWangA2019Pavement}
\bibinfo{author}{B, W. W.~A.}, \bibinfo{author}{A, M.~W.}, \bibinfo{author}{A, H.~L.}, \bibinfo{author}{A, H.~Z.}, \bibinfo{author}{B, K.~W.}, \bibinfo{author}{C, C.~H.}, \bibinfo{author}{D, J.~W.}, \bibinfo{author}{E, S.~Z.}, \& \bibinfo{author}{E, J.~C.} (\bibinfo{year}{2019}).
\newblock \bibinfo{title}{Pavement crack image acquisition methods and crack extraction algorithms: A review}.
\newblock {\it \bibinfo{journal}{Journal of Traffic and Transportation Engineering (English Edition)}\/},  {\it \bibinfo{volume}{6}\/}, \bibinfo{pages}{535--556}.
\bibitem[{Benz \& Rodehorst(2021)}]{Benz2021Vision}
\bibinfo{author}{Benz, C.}, \& \bibinfo{author}{Rodehorst, V.} (\bibinfo{year}{2021}).
\newblock \bibinfo{title}{Model-based crack width estimation using rectangle transform}.
\newblock In {\it \bibinfo{booktitle}{2021 17th International Conference on Machine Vision and Applications (MVA)}\/} (pp. \bibinfo{pages}{1--5}).
\bibitem[{David~Jenkins et~al.(2018)David~Jenkins, Carr, Iglesias, Buggy \& Morison}]{David2018Semantic}
\bibinfo{author}{David~Jenkins, M.}, \bibinfo{author}{Carr, T.~A.}, \bibinfo{author}{Iglesias, M.~I.}, \bibinfo{author}{Buggy, T.}, \& \bibinfo{author}{Morison, G.} (\bibinfo{year}{2018}).
\newblock \bibinfo{title}{A deep convolutional neural network for semantic pixel-wise segmentation of road and pavement surface cracks}.
\newblock In {\it \bibinfo{booktitle}{2018 26th European Signal Processing Conference (EUSIPCO)}\/} (pp. \bibinfo{pages}{2120--2124}).
\bibitem[{Dung \& Anh(2019)}]{DUNG-fcnn-automatinon-19}
\bibinfo{author}{Dung, C.~V.}, \& \bibinfo{author}{Anh, L.~D.} (\bibinfo{year}{2019}).
\newblock \bibinfo{title}{Autonomous concrete crack detection using deep fully convolutional neural network}.
\newblock {\it \bibinfo{journal}{Automation in Construction}\/},  {\it \bibinfo{volume}{99}\/}, \bibinfo{pages}{52--58}.
\bibitem[{Han et~al.(2024)Han, Shin \& Paik}]{Jihoon2024Robust}
\bibinfo{author}{Han, J.}, \bibinfo{author}{Shin, M.}, \& \bibinfo{author}{Paik, J.} (\bibinfo{year}{2024}).
\newblock \bibinfo{title}{Robust point cloud registration using hough voting-based correspondence outlier rejection}.
\newblock {\it \bibinfo{journal}{Engineering Applications of Artificial Intelligence}\/},  {\it \bibinfo{volume}{133}\/}, \bibinfo{pages}{107985}.
\bibitem[{J(1969)}]{skeletons-1969-Hilditch}
\bibinfo{author}{J, H.~C.} (\bibinfo{year}{1969}).
\newblock \bibinfo{title}{Linear skeletons from square cupboards}.
\newblock {\it \bibinfo{journal}{Machine intelligence}\/},  {\it \bibinfo{volume}{6}\/}, \bibinfo{pages}{403--420}.
\bibitem[{K(1901)}]{PCA_1901_PearsonK}
\bibinfo{author}{K, P.} (\bibinfo{year}{1901}).
\newblock \bibinfo{title}{Liii. on lines and planes of closest fit to systems of points in space}.
\newblock {\it \bibinfo{journal}{The London, Edinburgh, and Dublin Philosophical Magazine and Journal of Science}\/},  {\it \bibinfo{volume}{2}\/}, \bibinfo{pages}{559--572}.
\bibitem[{Kim \& Cho(2019)}]{Kim2019Image-based}
\bibinfo{author}{Kim, B.}, \& \bibinfo{author}{Cho, S.} (\bibinfo{year}{2019}).
\newblock \bibinfo{title}{Image-based concrete crack assessment using mask and region-based convolutional neural network}.
\newblock {\it \bibinfo{journal}{Structural Control and Health Monitoring}\/},  {\it \bibinfo{volume}{26}\/}, \bibinfo{pages}{e2381}.
\newblock \bibinfo{note}{E2381 STC-18-0418.R2}.
\bibitem[{L et~al.(2016)L, F, D \& J}]{dcnn-2016-zhang}
\bibinfo{author}{L, Z.}, \bibinfo{author}{F, Y.}, \bibinfo{author}{D, Z.~Y.}, \& \bibinfo{author}{J, Z.~Y.} (\bibinfo{year}{2016}).
\newblock \bibinfo{title}{Road crack detection using deep convolutional neural network}.
\newblock In {\it \bibinfo{booktitle}{2016 IEEE international conference on image processing (ICIP)}\/} (pp. \bibinfo{pages}{3708--3712}).
\newblock \bibinfo{address}{Phoenix, AZ, USA}: \bibinfo{publisher}{IEEE}.
\bibitem[{Liu et~al.(2021{\natexlab{a}})Liu, Miao, Mertz, Xu \& Kong}]{liu2021crackformer}
\bibinfo{author}{Liu, H.}, \bibinfo{author}{Miao, X.}, \bibinfo{author}{Mertz, C.}, \bibinfo{author}{Xu, C.}, \& \bibinfo{author}{Kong, H.} (\bibinfo{year}{2021}{\natexlab{a}}).
\newblock \bibinfo{title}{Crackformer: Transformer network for fine-grained crack detection}.
\newblock In {\it \bibinfo{booktitle}{Proceedings of the IEEE/CVF international conference on computer vision}\/} (pp. \bibinfo{pages}{3783--3792}).
\bibitem[{Liu et~al.(2019)Liu, Yao, Lu, Xie \& Li}]{liu2019deepcrack}
\bibinfo{author}{Liu, Y.}, \bibinfo{author}{Yao, J.}, \bibinfo{author}{Lu, X.}, \bibinfo{author}{Xie, R.}, \& \bibinfo{author}{Li, L.} (\bibinfo{year}{2019}).
\newblock \bibinfo{title}{Deepcrack: A deep hierarchical feature learning architecture for crack segmentation}.
\newblock {\it \bibinfo{journal}{Neurocomputing}\/},  {\it \bibinfo{volume}{338}\/}, \bibinfo{pages}{139--153}.
\bibitem[{Liu et~al.(2021{\natexlab{b}})Liu, Lin, Cao, Hu, Wei, Zhang, Lin \& Guo}]{liu2021swin}
\bibinfo{author}{Liu, Z.}, \bibinfo{author}{Lin, Y.}, \bibinfo{author}{Cao, Y.}, \bibinfo{author}{Hu, H.}, \bibinfo{author}{Wei, Y.}, \bibinfo{author}{Zhang, Z.}, \bibinfo{author}{Lin, S.}, \& \bibinfo{author}{Guo, B.} (\bibinfo{year}{2021}{\natexlab{b}}).
\newblock \bibinfo{title}{Swin transformer: Hierarchical vision transformer using shifted windows}.
\newblock In {\it \bibinfo{booktitle}{Proceedings of the IEEE/CVF international conference on computer vision}\/} (pp. \bibinfo{pages}{10012--10022}).
\bibitem[{Ni et~al.(2019)Ni, Zhang \& Chen}]{Ni2019Zernike}
\bibinfo{author}{Ni, F.~T.}, \bibinfo{author}{Zhang, J.}, \& \bibinfo{author}{Chen, Z.~Q.} (\bibinfo{year}{2019}).
\newblock \bibinfo{title}{Zernike-moment measurement of thin-crack width in images enabled by dual-scale deep learning}.
\newblock {\it \bibinfo{journal}{Computer-Aided Civil and Infrastructure Engineering}\/},  {\it \bibinfo{volume}{34}\/}, \bibinfo{pages}{367--384}.
\bibitem[{Oliveira \& Correia(2012)}]{AutomaticDetectionChara-2012-Oliveira}
\bibinfo{author}{Oliveira, H.}, \& \bibinfo{author}{Correia, P.} (\bibinfo{year}{2012}).
\newblock \bibinfo{title}{Automatic road crack detection and characterization}.
\newblock {\it \bibinfo{journal}{IEEE Transactions on Intelligent Transportation Systems}\/},  {\it \bibinfo{volume}{14}\/}, \bibinfo{pages}{155--168}.
\bibitem[{Ong et~al.(2022)Ong, Ismadi \& Wang}]{Ong2022hybridmethod}
\bibinfo{author}{Ong, J. C.~H.}, \bibinfo{author}{Ismadi, M.-Z.~P.}, \& \bibinfo{author}{Wang, X.} (\bibinfo{year}{2022}).
\newblock \bibinfo{title}{A hybrid method for pavement crack width measurement}.
\newblock {\it \bibinfo{journal}{Measurement}\/},  (p. \bibinfo{pages}{197}).
\bibitem[{Pang et~al.(2024)Pang, Xiong \& Wu}]{DBLP:journals-pang-its-2024}
\bibinfo{author}{Pang, J.}, \bibinfo{author}{Xiong, B.}, \& \bibinfo{author}{Wu, J.} (\bibinfo{year}{2024}).
\newblock \bibinfo{title}{Modeling multi-granularity context information flow for pavement crack detection}.
\newblock {\it \bibinfo{journal}{CoRR}\/},  {\it \bibinfo{volume}{abs/2404.12702}\/}. \href{http://arxiv.org/abs/2404.12702}{\tt arXiv:2404.12702}.
\bibitem[{Payab et~al.(2019)Payab, Abbasina \& Khanzadi}]{Payab2019Review}
\bibinfo{author}{Payab, M.}, \bibinfo{author}{Abbasina, R.}, \& \bibinfo{author}{Khanzadi, M.} (\bibinfo{year}{2019}).
\newblock \bibinfo{title}{A brief review and a new graph-based image analysis for concrete crack quantification}.
\newblock {\it \bibinfo{journal}{Springer Netherlands}\/}, .
\bibitem[{Qin et~al.(2018)Qin, Zou, Zheng, Zhang, Qingquan, Li, Xianbiao, Qi, Qian \& Wang}]{Qin2018DeepCrack}
\bibinfo{author}{Qin}, \bibinfo{author}{Zou}, \bibinfo{author}{Zheng}, \bibinfo{author}{Zhang}, \bibinfo{author}{Qingquan}, \bibinfo{author}{Li}, \bibinfo{author}{Xianbiao}, \bibinfo{author}{Qi}, \bibinfo{author}{Qian}, \& \bibinfo{author}{Wang} (\bibinfo{year}{2018}).
\newblock \bibinfo{title}{Deepcrack: Learning hierarchical convolutional features for crack detection}.
\newblock {\it \bibinfo{journal}{IEEE Transactions on Image Processing A Publication of the IEEE Signal Processing Society}\/}, .
\bibitem[{Qu et~al.(2021)Qu, Cao, Liu \& Zhou}]{Qu2021Pavement}
\bibinfo{author}{Qu, Z.}, \bibinfo{author}{Cao, C.}, \bibinfo{author}{Liu, L.}, \& \bibinfo{author}{Zhou, D.~Y.} (\bibinfo{year}{2021}).
\newblock \bibinfo{title}{A deeply supervised convolutional neural network for pavement crack detection with multiscale feature fusion}.
\newblock {\it \bibinfo{journal}{IEEE Transactions on Neural Networks and Learning Systems}\/},  {\it \bibinfo{volume}{PP}\/}, \bibinfo{pages}{1--10}.
\bibitem[{Raza \& {Arsalan Khushnood}(2022)}]{raza-22-consruction}
\bibinfo{author}{Raza, A.}, \& \bibinfo{author}{{Arsalan Khushnood}, R.} (\bibinfo{year}{2022}).
\newblock \bibinfo{title}{Digital image processing for precise evaluation of concrete crack repair using bio-inspired strategies}.
\newblock {\it \bibinfo{journal}{Construction and Building Materials}\/},  {\it \bibinfo{volume}{350}\/}, \bibinfo{pages}{128863}.
\bibitem[{Ronneberger et~al.(2015)Ronneberger, Fischer \& Brox}]{Ronneberger2015Biomedical}
\bibinfo{author}{Ronneberger, O.}, \bibinfo{author}{Fischer, P.}, \& \bibinfo{author}{Brox, T.} (\bibinfo{year}{2015}).
\newblock \bibinfo{title}{U-net: Convolutional networks for biomedical image segmentation}.
\newblock In {\it \bibinfo{booktitle}{International Conference on Medical Image Computing and Computer-Assisted Intervention}\/}.
\bibitem[{S \& I(2004)}]{AunifyingFramework-2004-Baker}
\bibinfo{author}{S, B.}, \& \bibinfo{author}{I, M.} (\bibinfo{year}{2004}).
\newblock \bibinfo{title}{A unifying framework}.
\newblock {\it \bibinfo{journal}{International journal of computer vision}\/},  {\it \bibinfo{volume}{56}\/}, \bibinfo{pages}{221--255}.
\bibitem[{Shi et~al.(2016{\natexlab{a}})Shi, Cui, Qi \& Meng}]{RandomStructuredForests-2016-YongShi}
\bibinfo{author}{Shi, Y.}, \bibinfo{author}{Cui, L.}, \bibinfo{author}{Qi, Z.}, \& \bibinfo{author}{Meng, F.} (\bibinfo{year}{2016}{\natexlab{a}}).
\newblock \bibinfo{title}{Automatic road crack detection using random structured forests}.
\newblock {\it \bibinfo{journal}{IEEE Transactions on Intelligent Transportation Systems}\/},  {\it \bibinfo{volume}{17}\/}, \bibinfo{pages}{3434--3445}.
\bibitem[{Shi et~al.(2016{\natexlab{b}})Shi, Cui, Qi, Meng \& Chen}]{shi2016automatic}
\bibinfo{author}{Shi, Y.}, \bibinfo{author}{Cui, L.}, \bibinfo{author}{Qi, Z.}, \bibinfo{author}{Meng, F.}, \& \bibinfo{author}{Chen, Z.} (\bibinfo{year}{2016}{\natexlab{b}}).
\newblock \bibinfo{title}{Automatic road crack detection using random structured forests}.
\newblock {\it \bibinfo{journal}{IEEE Transactions on Intelligent Transportation Systems}\/},  {\it \bibinfo{volume}{17}\/}, \bibinfo{pages}{3434--3445}.
\bibitem[{Song et~al.(2020)Song, Jia, Zhu, Jia, Gao et~al.}]{Song2020automated}
\bibinfo{author}{Song, W.}, \bibinfo{author}{Jia, G.}, \bibinfo{author}{Zhu, H.}, \bibinfo{author}{Jia, D.}, \bibinfo{author}{Gao, L.} et~al. (\bibinfo{year}{2020}).
\newblock \bibinfo{title}{Automated pavement crack damage detection using deep multiscale convolutional features}.
\newblock {\it \bibinfo{journal}{Journal of Advanced Transportation}\/},  {\it \bibinfo{volume}{2020}\/}.
\bibitem[{Takafumi et~al.(2011)Takafumi, Nishikawa, Junji, Yoshida, Toshiyuki, Sugiyama, Yozo \& Fujino}]{Takafumi2011Concrete}
\bibinfo{author}{Takafumi}, \bibinfo{author}{Nishikawa}, \bibinfo{author}{Junji}, \bibinfo{author}{Yoshida}, \bibinfo{author}{Toshiyuki}, \bibinfo{author}{Sugiyama}, \bibinfo{author}{Yozo}, \& \bibinfo{author}{Fujino} (\bibinfo{year}{2011}).
\newblock \bibinfo{title}{Concrete crack detection by multiple sequential image filtering}.
\newblock {\it \bibinfo{journal}{Computer-Aided Civil and Infrastructure Engineering}\/}, .
\bibitem[{Wang et~al.(2017)Wang, Kelvin, C., P., Shaofan, Qiu, Shi \& Wenjuan}]{Wang2017Methodology}
\bibinfo{author}{Wang}, \bibinfo{author}{Kelvin}, \bibinfo{author}{C.}, \bibinfo{author}{P.}, \bibinfo{author}{Shaofan}, \bibinfo{author}{Qiu}, \bibinfo{author}{Shi}, \& \bibinfo{author}{Wenjuan} (\bibinfo{year}{2017}).
\newblock \bibinfo{title}{Methodology for accurate aashto pp67-10-based cracking quantification using 1-mm 3d pavement images}.
\newblock {\it \bibinfo{journal}{Journal of Computing in Civil Engineering}\/}, .
\bibitem[{Wang et~al.(2018)Wang, Zhang, Wang, Braham \& Qiu}]{Wang2018PavementCrack}
\bibinfo{author}{Wang, W.}, \bibinfo{author}{Zhang, A.}, \bibinfo{author}{Wang, K. C.~P.}, \bibinfo{author}{Braham, A.~F.}, \& \bibinfo{author}{Qiu, S.} (\bibinfo{year}{2018}).
\newblock \bibinfo{title}{Pavement crack width measurement based on laplace's equation for continuity and unambiguity}.
\newblock {\it \bibinfo{journal}{Computer‐Aided Civil and Infrastructure Engineering}\/}, .
\bibitem[{Weng et~al.(2019)Weng, Huang \& Wang}]{Weng2019Segment}
\bibinfo{author}{Weng, X.}, \bibinfo{author}{Huang, Y.}, \& \bibinfo{author}{Wang, W.} (\bibinfo{year}{2019}).
\newblock \bibinfo{title}{Segment-based pavement crack quantification}.
\newblock {\it \bibinfo{journal}{Automation in construction}\/},  {\it \bibinfo{volume}{105}\/}, \bibinfo{pages}{102819.1--102819.16}.
\bibitem[{Xiang et~al.(2022)Xiang, Wang, Deng, Shi \& Kong}]{Xiang2022concrete}
\bibinfo{author}{Xiang, C.}, \bibinfo{author}{Wang, W.}, \bibinfo{author}{Deng, L.}, \bibinfo{author}{Shi, P.}, \& \bibinfo{author}{Kong, X.} (\bibinfo{year}{2022}).
\newblock \bibinfo{title}{Crack detection algorithm for concrete structures based on super-resolution reconstruction and segmentation network}.
\newblock {\it \bibinfo{journal}{Automation in construction}\/},  (p. \bibinfo{pages}{140}).
\bibitem[{Xu et~al.(2023{\natexlab{a}})Xu, Yue \& Liu}]{XU-eaai-crack-segmentation-23}
\bibinfo{author}{Xu, G.}, \bibinfo{author}{Yue, Q.}, \& \bibinfo{author}{Liu, X.} (\bibinfo{year}{2023}{\natexlab{a}}).
\newblock \bibinfo{title}{Deep learning algorithm for real-time automatic crack detection, segmentation, qualification}.
\newblock {\it \bibinfo{journal}{Engineering Applications of Artificial Intelligence}\/},  {\it \bibinfo{volume}{126}\/}, \bibinfo{pages}{107085}.
\bibitem[{Xu et~al.(2023{\natexlab{b}})Xu, Yue \& Liu}]{XU2023real-time}
\bibinfo{author}{Xu, G.}, \bibinfo{author}{Yue, Q.}, \& \bibinfo{author}{Liu, X.} (\bibinfo{year}{2023}{\natexlab{b}}).
\newblock \bibinfo{title}{Deep learning algorithm for real-time automatic crack detection, segmentation, qualification}.
\newblock {\it \bibinfo{journal}{Engineering Applications of Artificial Intelligence}\/},  {\it \bibinfo{volume}{126}\/}, \bibinfo{pages}{107085}.
\bibitem[{Yamaguchi \& Hashimoto(2010)}]{Yamaguchi2010Practical}
\bibinfo{author}{Yamaguchi, T.}, \& \bibinfo{author}{Hashimoto, S.} (\bibinfo{year}{2010}).
\newblock \bibinfo{title}{Practical image measurement of crack width for real concrete structure}.
\newblock {\it \bibinfo{journal}{Electronics Communications in Japan}\/},  {\it \bibinfo{volume}{92}\/}, \bibinfo{pages}{1--12}.
\bibitem[{Yang et~al.(2020)Yang, Zhang, Yu, Prokhorov, Mei \& Ling}]{yan-crack500-its-2020}
\bibinfo{author}{Yang, F.}, \bibinfo{author}{Zhang, L.}, \bibinfo{author}{Yu, S.}, \bibinfo{author}{Prokhorov, D.}, \bibinfo{author}{Mei, X.}, \& \bibinfo{author}{Ling, H.} (\bibinfo{year}{2020}).
\newblock \bibinfo{title}{Feature pyramid and hierarchical boosting network for pavement crack detection}.
\newblock {\it \bibinfo{journal}{IEEE Transactions on Intelligent Transportation Systems}\/},  {\it \bibinfo{volume}{21}\/}, \bibinfo{pages}{1525--1535}.
\bibitem[{Yang et~al.(2017)Yang, Liu, Song \& Li}]{Yang2017Estimation}
\bibinfo{author}{Yang, J.}, \bibinfo{author}{Liu, X.}, \bibinfo{author}{Song, X.}, \& \bibinfo{author}{Li, K.} (\bibinfo{year}{2017}).
\newblock \bibinfo{title}{Estimation of signal-dependent noise level function using multi-column convolutional neural network}.
\newblock In {\it \bibinfo{booktitle}{IEEE International Conference on Image Processing}\/}.
\bibitem[{Z et~al.(2012)Z, A \& et~al.}]{TITL-2012-Zhang}
\bibinfo{author}{Z, Z.}, \bibinfo{author}{A, G.}, \& \bibinfo{author}{et~al., L.~X.} (\bibinfo{year}{2012}).
\newblock \bibinfo{title}{Tilt: Transform invariant low-rank textures}.
\newblock {\it \bibinfo{journal}{International journal of computer vision}\/},  {\it \bibinfo{volume}{99}\/}, \bibinfo{pages}{1--24}.
\bibitem[{Zhang et~al.(2010)Zhang, Ganesh, Liang \& Ma}]{Zhang2010TILT}
\bibinfo{author}{Zhang, Z.}, \bibinfo{author}{Ganesh, A.}, \bibinfo{author}{Liang, X.}, \& \bibinfo{author}{Ma, Y.} (\bibinfo{year}{2010}).
\newblock \bibinfo{title}{Tilt: Transform invariant low-rank textures}.
\newblock {\it \bibinfo{journal}{International Journal of Computer Vision}\/},  {\it \bibinfo{volume}{99}\/}, \bibinfo{pages}{1--24}.
\bibitem[{Zhou \& Song(2020)}]{ZHOU-autiomation-construction-20}
\bibinfo{author}{Zhou, S.}, \& \bibinfo{author}{Song, W.} (\bibinfo{year}{2020}).
\newblock \bibinfo{title}{Concrete roadway crack segmentation using encoder-decoder networks with range images}.
\newblock {\it \bibinfo{journal}{Automation in Construction}\/},  {\it \bibinfo{volume}{120}\/}, \bibinfo{pages}{103403}.
\bibitem[{Zou et~al.(2012)Zou, Cao, Li, Mao \& Wang}]{zou2012cracktree}
\bibinfo{author}{Zou, Q.}, \bibinfo{author}{Cao, Y.}, \bibinfo{author}{Li, Q.}, \bibinfo{author}{Mao, Q.}, \& \bibinfo{author}{Wang, S.} (\bibinfo{year}{2012}).
\newblock \bibinfo{title}{Cracktree: Automatic crack detection from pavement images}.
\newblock {\it \bibinfo{journal}{Pattern Recognition Letters}\/},  {\it \bibinfo{volume}{33}\/}, \bibinfo{pages}{227--238}.
\bibitem[{Zou et~al.(2019)Zou, Zhang, Li, Qi, Wang \& Wang}]{zhou-tip-crack-segmentation-2019}
\bibinfo{author}{Zou, Q.}, \bibinfo{author}{Zhang, Z.}, \bibinfo{author}{Li, Q.}, \bibinfo{author}{Qi, X.}, \bibinfo{author}{Wang, Q.}, \& \bibinfo{author}{Wang, S.} (\bibinfo{year}{2019}).
\newblock \bibinfo{title}{Deepcrack: Learning hierarchical convolutional features for crack detection}.
\newblock {\it \bibinfo{journal}{IEEE Transactions on Image Processing}\/},  {\it \bibinfo{volume}{28}\/}, \bibinfo{pages}{1498--1512}.

\end{thebibliography}

%

\end{document}